\documentclass{article}

\usepackage{arxiv}

\usepackage[utf8]{inputenc} % allow utf-8 input
\usepackage[T1]{fontenc}    % use 8-bit T1 fonts

\usepackage{hyperref}
\hypersetup{hidelinks}

\usepackage{url}            % simple URL typesetting
\usepackage{booktabs}       % professional-quality tables
\usepackage{amsfonts}       % blackboard math symbols
\usepackage{nicefrac}       % compact symbols for 1/2, etc.
\usepackage{microtype}      % microtypography
\usepackage{lipsum}
\usepackage{graphicx}
\usepackage{multirow}
\usepackage[numbers,square]{natbib}
\usepackage{doi}
\usepackage{amsmath}
\usepackage[bottom]{footmisc}
\usepackage{footnote}
\usepackage{natbib}
\usepackage{lipsum}
\usepackage{algorithm}
\usepackage{algpseudocode}
\usepackage[inline,shortlabels]{enumitem}
\usepackage{subcaption}
\usepackage{amssymb}
\usepackage{textcomp}
\usepackage{xcolor}
\usepackage{makecell}
\usepackage{float}
\usepackage{arydshln}

\title{Semantic Segmentation Refiner for Ultrasound Applications with Zero-Shot Foundation Models}

\date{}


\author{\href{https://orcid.org/0000-0003-3741-9166}{\includegraphics[scale=0.06]{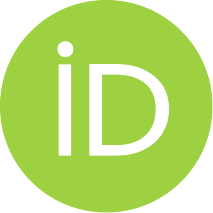}\hspace{1mm}Hedda Cohen Indelman$^{1, \dag}$} \\
    AI/ML Research, GE Healthcare\\
		\And
  \href{https://orcid.org/0009-0009-5635-6763}{\includegraphics[scale=0.06]{orcid.pdf}\hspace{1mm}Elay Dahan$^{1, \dag}$} \\
	AI/ML Research, GE Healthcare\\
		\And
  \href{https://orcid.org/0009-0009-0370-5448}{\includegraphics[scale=0.06]{orcid.pdf}\hspace{1mm}Angeles M. Perez-Agosto$^{2}$}\\
    Clinical Insights \& Development, GE Healthcare\\
    		\And
  \href{https://orcid.org/0009-0000-5542-0035}{\includegraphics[scale=0.06]{orcid.pdf}\hspace{1mm}Carmit Shiran$^{3}$}\\
    Clinical Insights \& Development, GE Healthcare\\
	    \And
     \href{https://orcid.org/0009-0009-0263-6102}{\includegraphics[scale=0.06]{orcid.pdf}\hspace{1mm}Doron Shaked$^{1}$} \\
	AI/ML Research, GE Healthcare\\
        \And
     \href{https://orcid.org/0000-0002-0939-3379}{\includegraphics[scale=0.06]{orcid.pdf}\hspace{1mm}Nati Daniel$^{1,}$\thanks{Corresponding author, e-mail: \href{mailto:nati.daniel@gehealthcare.com}{nati.daniel@gehealthcare.com}. $\dag$These authors have contributed equally to this work. $^{1}$Dept. of AI/ML Research, GE Healthcare, Haifa, Israel. $^{2}$Dept. of Clinical Applications, Point of Care Ultrasound \& Handheld, Texas, USA. $^{3}$Dept. of Clinical Applications, Point of Care Ultrasound \& Handheld, Wisconsin, USA.
     }} \\
	AI/ML Research, GE Healthcare
}

\renewcommand{\headeright}{Cohen H et al.}
\renewcommand{\shorttitle}{Semantic Segmentation Refiner for U/S Applications with Zero-Shot FMs}

\begin{document}

\maketitle

\begin{abstract}
Despite the remarkable success of deep learning in medical imaging analysis, medical image segmentation remains challenging due to the scarcity of high-quality labeled images for supervision. Further, the significant domain gap between natural and medical images in general and ultrasound images in particular hinders fine-tuning models trained on natural images to the task at hand. In this work, we address the performance degradation of segmentation models in low-data regimes and propose a prompt-less segmentation method harnessing the ability of segmentation foundation models to segment abstract shapes. We do that via our novel prompt point generation algorithm which uses coarse semantic segmentation masks as input and a zero-shot prompt-able foundation model as an optimization target. We demonstrate our method on a segmentation findings task (pathologic anomalies) in ultrasound images. Our method's advantages are brought to light in varying degrees of low-data regime experiments on a small-scale musculoskeletal ultrasound images dataset, yielding a larger performance gain as the training set size decreases.
\end{abstract}


\section{Introduction}\label{intro}

\begin{figure*}[ht!]
    \centering
    \includegraphics[width=\linewidth]{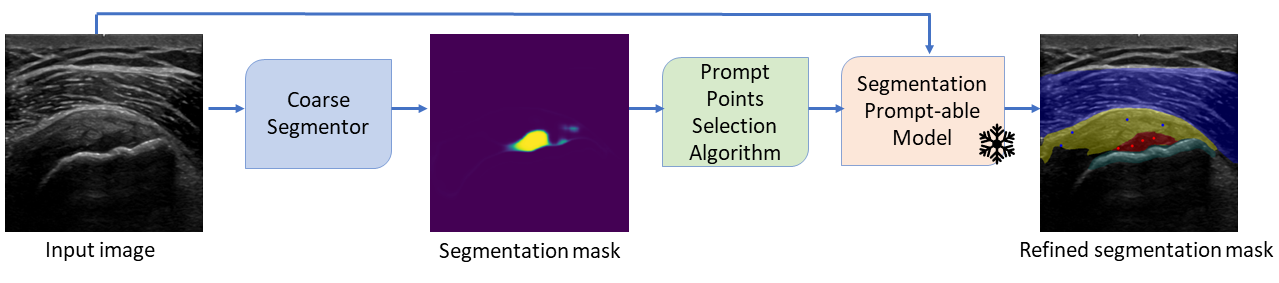}
    \caption{A high-level illustration of our semantic segmentation refinement method with zero-shot foundation models. A pre-trained segmentation model predicts a semantic segmentation for each class of an input image. In this example, classes comprise anatomies and pathologies in an ultrasound image, and the coarse segmentor output depicts the predicted semantic segmentation of a pathology. A prompt selection model selects positive and negative points. Consequently, a zero-shot semantic segmentation mask of the pathology is predicted by a foundation segmentation model, prompted by the selected points for the input image. Positive prompt points are depicted in red, and negative prompt points are depicted in blue. The pathology semantic segmentation prediction is highlighted in red. For illustration purposes, the muscle is highlighted in purple, the tendon in yellow, and the bone in green. The freeze symbol indicates preventing gradients from being propagated to the model weights.}
    \label{fig:high_level_architecture}
\end{figure*}

\iffalse Musculoskeletal (MSK) \cite{parker2008musculoskeletal} \fi Ultrasound is a popular medical imaging modality used to image a large variety of organs and tissues. Ultrasound is often the preferred choice due to its non-radiative and non-invasive nature, relatively easy and fast imaging procedure, and lower costs. Automating the diagnosis or highlighting relevant areas in the image will contribute to faster workflows and potentially more consistent and accurate diagnoses. 

Artificial Intelligence (AI) has demonstrated remarkable success in automatic medical imaging analysis. Compared to classical methods, previous work based on convolutional neural networks on various medical imaging tasks, such as classification and segmentation, have shown state-of-the-art results \cite{liu2019deep, 10.1007/978-3-030-00937-3_16, pmid33828217, Topol2019}. However, effective deep learning segmentation
algorithms for medical images is an especially challenging task due to the scarcity of high-quality labeled images for supervision. Moreover, in medical imaging it is often the case that identification of \textit{findings} regions, namely regions of potentially pathological visual anomalies, having neither a clear boundary nor a typical geometry or position is much more challenging than the identification of an anatomy in its context. Findings are also typically rare, which brings to light the challenge of training such models in limited data regimes.

Recently, new segmentation models have emerged. Trained on data at huge scales, these foundation models aim to be more generic rather than tailored to specific datasets. The Segment Anything Model (SAM) \cite{kirillov2023segment} is a foundational model demonstrating zero-shot generalization in
segmenting natural images using a prompt-driven approach. The SonoSAM \cite{ravishankar2023sonosam} foundational model adapts SAM to ultrasound images by fine-tuning the prompt and mask decoder  \cite{ravishankar2023sonosam}. 
Although fine-tuning methods often improve the results on target datasets \cite{Huang_2024} they essentially downgrade the generalization capabilities of the foundation model. Further, a significant domain gap between natural and medical images, ultrasound images in particular\cite{cheng2023sammed2d}, hinders fine-tuning models trained on natural images to the task at hand \cite{Huang_2024}.

In this work, we address the performance degradation of segmentation models in low-data regimes and derive a novel method for harnessing segmentation foundation models' ability to segment arbitrary regions. 
Our semantic segmentation refinement method comprises two stages: First, a coarse segmentation is predicted by a model trained on a small subset of the training data.
In the second stage, our novel points generation from a coarse pathology segmentation algorithm is used to prompt a segmentation foundation model. Positive prompt points are selected using a partition around medoids method as the most representative pathology points. Negative prompt points are selected by a prompt selection optimization algorithm that identify the context anatomy.
Importantly, we do not fine-tune the foundation model to our dataset, i.e., it produces a zero-shot segmentation. The end-to-end pipeline is illustrated in Fig. \ref{fig:high_level_architecture}.

The method's advantages are brought to light on varying degrees of low-data regimes experiments on a small-scale images dataset, yielding a larger performance gain compared to a state-of-the-art segmentation model \cite{chen2017rethinking} as the training set size decreases. Further, ablation studies validate the effectiveness of our semantic segmentation refinement model. Our approach applies to other ultrasound-based medical diagnostics tasks.

The paper is organized as follows: Section \ref{related} presents the semantic segmentation task and leading approaches. 
Our method is presented in Section \ref{metod}, and the experimental setup is presented in Section \ref{experiments}. 
Section \ref{results} presents the results and ablation studies on a discontinuity in tendon fiber (DITF) pathology finding task in a musculoskeletal ultrasound (MSK) dataset, and the conclusions are presented in Section \ref{conclusions}.

\section{Related Work}\label{related}

\subsection{Semantic Segmentation Models}
Semantic segmentation aims to assign a label or a class to each pixel in an image. 
Unlike image classification, which assigns a single label to the entire image, semantic segmentation provides a more detailed understanding of the visual scene by segmenting it into distinct regions corresponding to objects or classes. This is an essential technique for applications, such as autonomous vehicles, medical image analysis, and scene understanding in robotics.

Like other computer vision tasks, deep learning has demonstrated state-of-the-art results in the semantic segmentation of medical images. The semantic segmentation problem can be formulated as follows:
Given an image $\mathbf{I}$ $\in$ $\mathbf{R}^{C \times H \times W}$, our goal is to train a deep neural network to predict the pixel-wise probability map $\mathbf{S}^{N \times H \times W}$ of the classes in the dataset, where $\mathbf{N}$ is the number of classes in the dataset.

DeepLabV3 \cite{chen2017rethinking} represents a distinctive approach in semantic image segmentation. Utilizing dilated convolutions, the model strategically enlarges the receptive field and manages the balance between global and local features through padding rates. Notably, the spatial pyramid pooling module proposed by the authors aggregates features from dilated convolutions at various scales, enhancing contextual information. Distinctive from encoder-decoder architectures such as the U-Net \cite{10.1007/978-3-319-24574-4_28}, it is built upon a robust pre-trained encoder, contributing to its success in generating accurate and detailed segmentation masks across diverse applications.

Since DeepLabV3 remains a staple choice for a performant semantic segmentation model, we adopt it as our method's coarse segmentor.  

\subsection{Semantic Segmentation Foundation Models}
Foundation models are trained on broad data at a huge scale and are adaptable to a wide range of downstream tasks \cite{devlin-etal-2019-bert, pmlr-v139-ramesh21a, NEURIPS2020_1457c0d6}.
The Segment Anything
Model (SAM) \cite{kirillov2023segment} emerged as a versatile foundation model for natural
image segmentation. Trained on a dataset of over 11 million images and 1B masks, it demonstrates impressive zero-shot generalization in segmenting natural images using an interactive and prompt-driven approach. Prompt types include foreground/background points, bounding boxes, masks, and text prompts. However, SAM achieves subpar generalization on medical images due to substantial domain gaps between natural and medical images \cite{he2023computervision,Mazurowski_2023, hu2023efficiently,ma2023segment, li2023autoprompting}.
Moreover, SAM obtains the poorest results on ultrasound compared to other medical imaging modalities \cite{Mazurowski_2023}. 
These results are attributed to the ultrasound
characteristics, e.g., the scan cone, poor image quality, and unique speckled texture.
A common methodology to overcome this generalization difficulty is to fine-tune a foundation model on a target dataset \cite{peng2023samparser}. 
An efficient fine-tuning strategy is Low-Rank Adaptation (LoRA) \cite{hu2022lora}, which has been adopted in fine-tuning SAM to relatively small medical imaging datasets
\cite{feng2023cheap, zhang2023customized, chen2023samocta}.  
SonoSAM \cite{ravishankar2023sonosam} demonstrates state-of-the-art generalization in segmenting ultrasound images. Fine-tuned on a rich and diverse set of ultrasound image-mask pairs, it has emerged as a prompt-able foundational model for ultrasound image segmentation. 

Notably, adapting prompt-based models to medical image segmentation is difficult due to the conundrum of crafting high-quality prompts \cite{Mazurowski_2023}.
Manually selecting prompts is time-consuming and requires domain expertise. Methods of extracting prompts from ground-truth masks \cite{chen2023samocta} cannot be applied during inference as they rely on full supervision. 
Auto-prompting techniques rely on the strong Vision Transformer (ViT-H) image encoder \cite{dosovitskiy2021image} semantic representation capabilities, and suggest generating a segmentation prompt based on SAM's image encoder embedding \cite{li2023autoprompting,anand2023oneshot}.
Other strategies suggest replacing the mask decoder with a prediction head requiring no prompts \cite{hu2023efficiently}.
Nevertheless, SAM's zero-shot prediction accuracy is typically lower than that of the segmentation models trained with fully supervised methods \cite{cheng2023sam}. 

Motivated by the generalization abilities of segmentation foundation models, we devise a points selection algorithm from coarse segmentation masks that allows harnessing prompt-based models to ultrasound segmentation in a zero-shot setting.

\section{Method}\label{metod}
In this section, we present our method for refining a coarse pathology segmentation mask with zero-shot foundation models. 
This method can be adapted to natural images, as well as to the medical imaging domain. Herein, we validate it based on a specific challenging task of segmenting a discontinuity of the tendon fiber finding (Sec. \ref{data_section}), which is the main ultrasound finding of a tendon partial tear pathology.

Our key intuition is that although the performance of segmentation models decreases significantly in low-data regimes, even such coarse segmentation masks can be utilized for extracting high-quality prompts that harness segmentation foundation models' capabilities. Importantly, we use the publicly available pre-trained foundation models without further modification. The flexibility of our method allows for incorporating either SonoSAM or SAM. Though the above-mentioned foundation models allow several types of prompts, we focus on foreground (positive) and background (negative) prompt points.

Our method makes use of the ground-truth tendon segmentation, denoted  $T^{gt}$. Since the tendon in the context of the DIFT pathology is usually easy to segment due to its typical geometry and position and relatively simple data acquisition and labeling, we assume that strong segmentation models exist for this task and that their output can be used in lieu of the ground-truth segmentation.   
With that, we introduce our two-stage method, summarized in Algorithm \ref{alg:method}.

\begin{algorithm}
\caption{The Semantic Segmentation Refiner Method} \label{alg:method}
\textbf{Input:} 
\begin{itemize}
    \item Input image $I$
    \item Ground-truth tendon mask $T^{gt}$ 
    \item Frozen $SonoSAM$ model
    \item Pre-trained segmentation model $S$
\end{itemize}
\textbf{Output:}
\begin{itemize}
    \item Refined pathology segmentation mask $O$
\end{itemize}
\begin{algorithmic}[1]
\State Coarse segmentation mask $\tilde O \gets$ $S(I)$
 
\State Positive points selection $pts^{pos} \gets$ $k$-medoids($\tilde O$)  
\State Modified ground-truth tendon mask $T^{\tilde{gt}} \gets T^{gt} \setminus \tilde O$
\State Initialize complementary problem \\
$\bar{pts}^{neg}  \gets pts^{pos} $, $\bar pts^{pos}  \gets \text{random from } T^{\tilde{gt}}$
    \For{$t$ \textbf{in} range($1,T$)}
        \State Optimize $\bar pts^{pos}$ as parameters: 
        \State $\ell_{ce}(\bar{pts}, T^{\tilde{gt}}) =  -T^{\tilde{gt}} \log \left( SonoSAM(I,\bar{pts})\right)$
        \State Update $\bar pts^{pos} \gets \bar pts^{pos}$
        \EndFor
    \State Flip: ${pts}^{neg}  \gets \bar pts^{pos}$
    \State Output $O \gets SonoSAM(I,{pts})$
\end{algorithmic}
\end{algorithm}
First, a segmentation model \cite{chen2017rethinking} is trained on a random subset of the training data. A coarse semantic segmentation is then predicted for a given test image.
Then, $k$ positive and $k$ negative prompt points are selected to prompt a segmentation
foundation model.
We next describe our prompt points selection algorithm in greater detail.

\subsection{Positive Points Selection}\label{positive_selection}
We aim to select points that are the most representative of the coarse pathology segmentation mask as the positive prompt points. This selection objective translates to the partitioning around the medoids method's approach. This approach is preferable compared to a selection based on a minimization of the sum of squared distance (i.e., the $k$-means) in the case of multiple pathology blobs since the latter might select centroids in between pathology blobs. Thus, $k$ mass centers of the coarse pathology segmentation masks are selected as positive points using the $k$- medoids clustering algorithm \cite{CAM1987}.

\begin{figure}[h]
    \centering
    \includegraphics[width=0.6\linewidth]{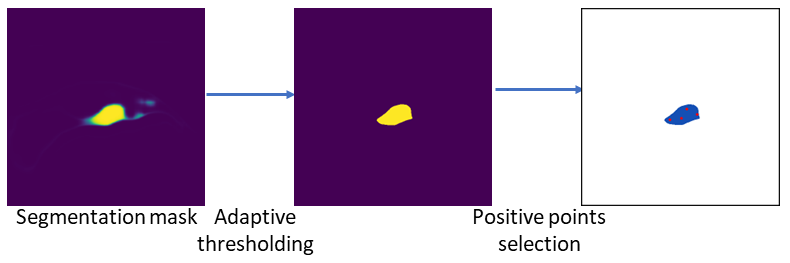}
    \caption{An illustration of our positive (foreground) points selection module, depicted in red. A threshold is applied to the coarse segmentation prediction. A $k$- medoids clustering algorithm is applied to select $k$ positive pathology points.}    \label{fig:positive_points_selection}
\end{figure}

To reduce the probability of selecting false positive points, a threshold is applied to the coarse pathology segmentation masks before selection. We denote the selected positive points as $pts^{pos} = \{pts^{pos}_i\}_{i=1}^{k}$. This process is illustrated in Fig. \ref{fig:positive_points_selection}.

\subsection{Negative Points Refinement}\label{negative_selection} 
We take inspiration from hard negative selection literature \cite{Robinson2020ContrastiveLW, Hafidi2021, zheng2023contrastive}, and aim to select the most informative negative points w.r.t. the foreground object. To that end, we formulate a complementary prompt points selection problem w.r.t. the background given the $k$ selected foreground points (\ref{positive_selection}),  $\bar{pts}= \{\bar{pts}^{pos}, \bar{pts}^{neg} \}$. When the foreground is the pathology, the background is the context anatomy, herein the background is a tendon anatomy.

The complementary prompt points selection is optimized to decrease the binary cross-entropy (BCE) loss between the foundation model's zero-shot tendon segmentation mask prompted on these points and a modified ground-truth tendon mask, denoted $T^{\tilde{gt}}$. To avoid predicting tendon points within foreground pathology, the values of the ground-truth tendon mask overlapping with the coarse pathology detection are modified to zero. 
As points initialization for this complementary problem, we flip the labels of $pts^{pos}$ such that they correspond to negative points, $\bar{pts}^{neg} \leftarrow pts^{pos} $. Further, $k$ points are selected at random from $T^{\tilde{gt}}$, denoted $\bar{pts}^{pos}$. While freezing the foundation model, the point prompt optimization is performed for a maximum of 100 steps or until convergence. The optimization is performed such that the selected points are optimal w.r.t. the complementary problem of the tendon segmentation given the foreground pathology predicted by the coarse segmentor.

\begin{figure}[htbp]
    \centering
    \includegraphics[width=0.75\linewidth]{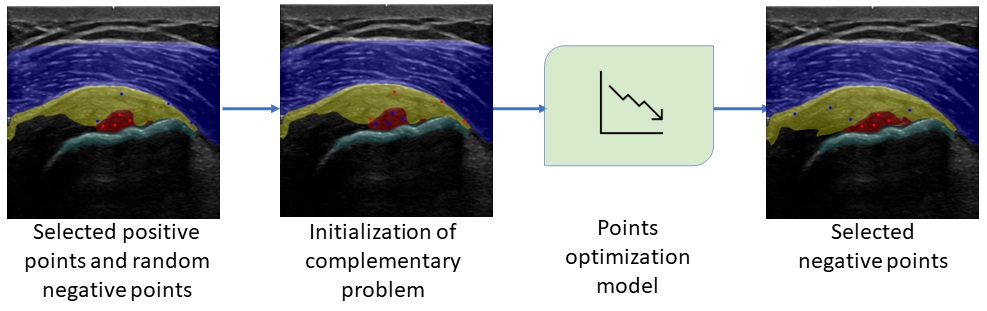}
    \caption{An illustration of our negative (background) points selection module. In addition to the positive selected points (Sec. \ref{positive_selection}), negative points are selected randomly from the modified ground-truth tendon mask.   
    The points are flipped to initialize the settings of the complementary tendon segmentation problem. Our points optimization model optimizes prompt points selection w.r.t. the complementary tendon zero-shot segmentation problem (Sec. \ref{negative_selection}). Finally, prompt points are again flipped to account for positive and negative prompt points towards the pathology segmentation.}
    \label{fig:negative_points_selection}
\end{figure}

Denote an input image as $I$, SonoSAM's zero-shot tendon segmentation given input $I$ and its corresponding optimized prompt points $\bar{pts}$ as $SonoSAM(I,\bar{pts})$. Then, the BCE loss of the complementary problem is:
\begin{equation}
    \ell_{ce}(\bar{pts}, T^{\tilde{gt}}) =  -T^{\tilde{gt}} \log \left( SonoSAM(I,\bar{pts}) \right).
\end{equation}
We used the AdamW \cite{loshchilov2017decoupled} optimizer, with learning rate of \(4e^{-3}\), and standard betas to optimize the positive points $\bar{pts}^{pos}$.
The optimized positive tendon points selected by this model serve as $k$ negative prompt points,  $pts^{neg} \leftarrow \bar{pts}^{pos}$, towards the foreground pathology segmentation. 
This process is illustrated in Fig. \ref{fig:negative_points_selection}.

\section{Experiments}\label{experiments}

\begin{figure*}[h!t]
\begin{minipage}{0.17\textwidth}
\includegraphics[width=\textwidth]{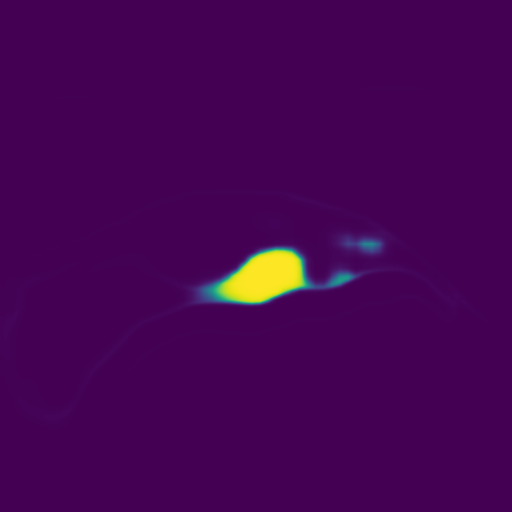}
\endminipage\hfill
\minipage{0.17\textwidth}
\includegraphics[width=\textwidth]{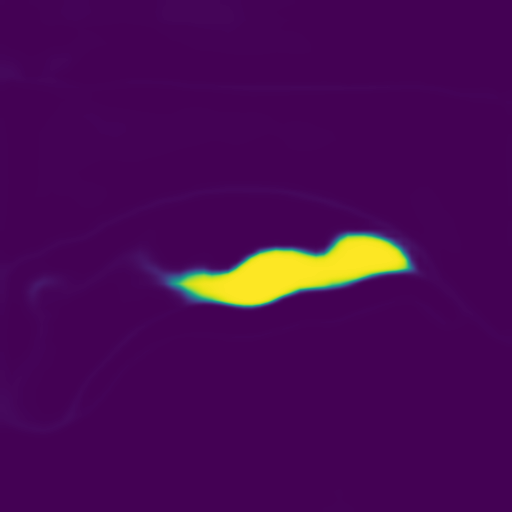}
\endminipage\hfill
\minipage{0.17\textwidth}
\includegraphics[width=\textwidth]{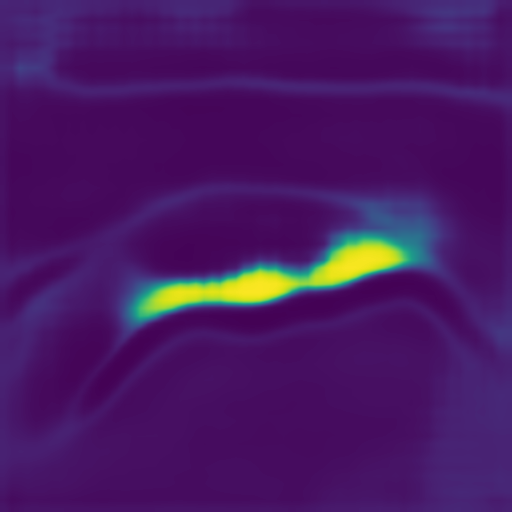}
\endminipage\hfill
\minipage{0.17\textwidth}
\includegraphics[width=\textwidth]{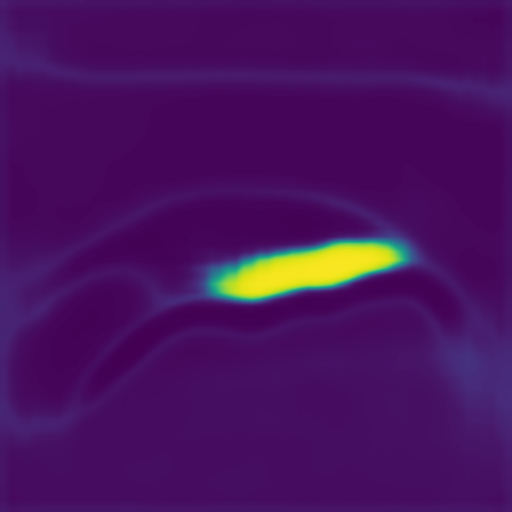}
\endminipage\hfill
\minipage{0.17\textwidth}
\includegraphics[width=\textwidth]{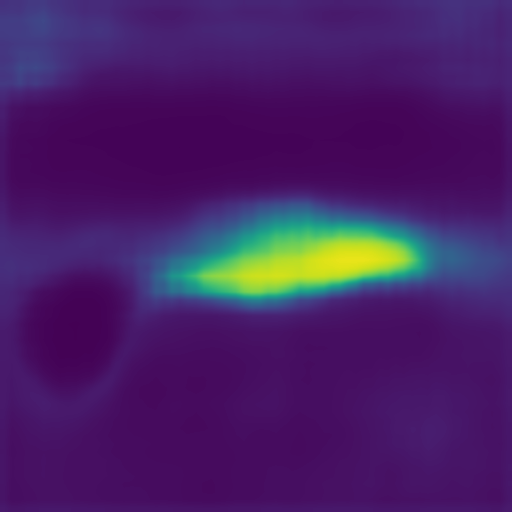}
\end{minipage}

\vspace{2pt}
\begin{minipage}{0.17\textwidth}
\includegraphics[width=\textwidth]{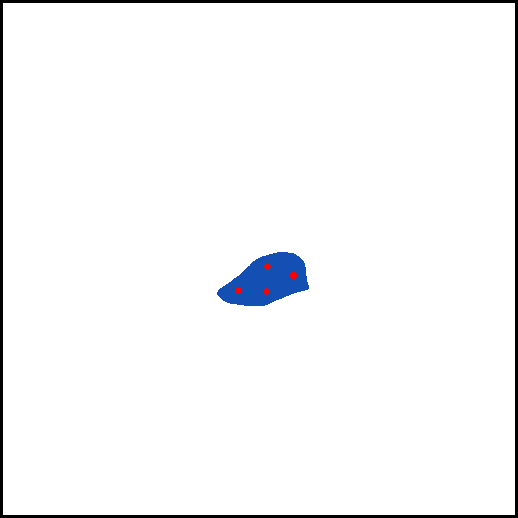}
  \label{fig:A100}
\endminipage\hfill
\minipage{0.17\textwidth}
\includegraphics[width=\textwidth]{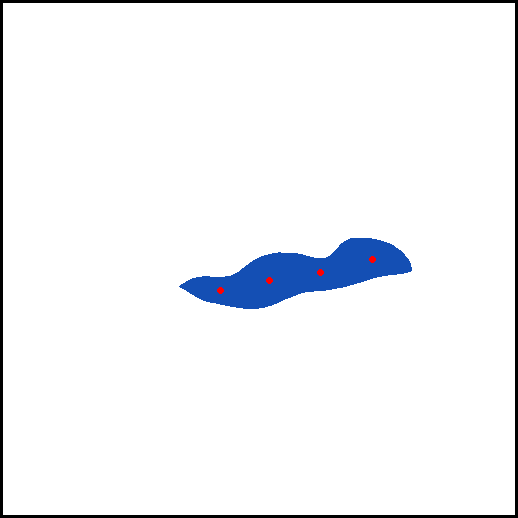}
    \label{fig:A35}
\endminipage\hfill
\minipage{0.17\textwidth}
\includegraphics[width=\textwidth]{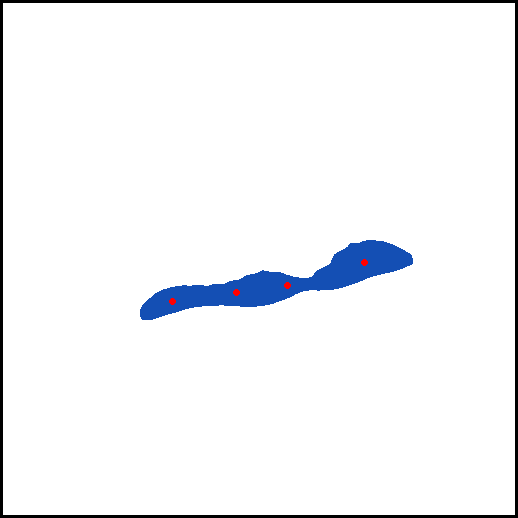}
    \label{fig:A15}
\endminipage\hfill
\minipage{0.17\textwidth}
\includegraphics[width=\textwidth]{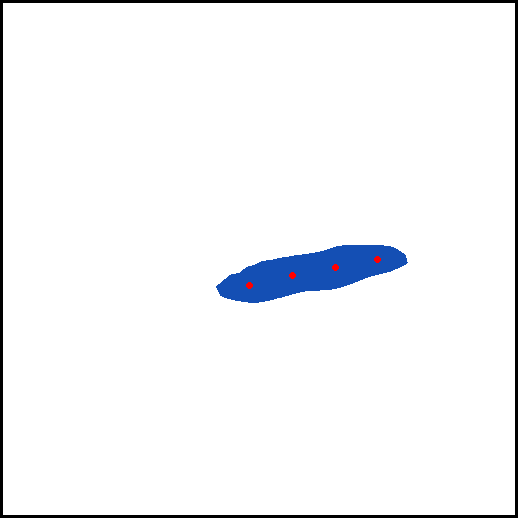}
  \label{fig:A8}
  \endminipage\hfill
\minipage{0.17\textwidth}
\includegraphics[width=\textwidth]{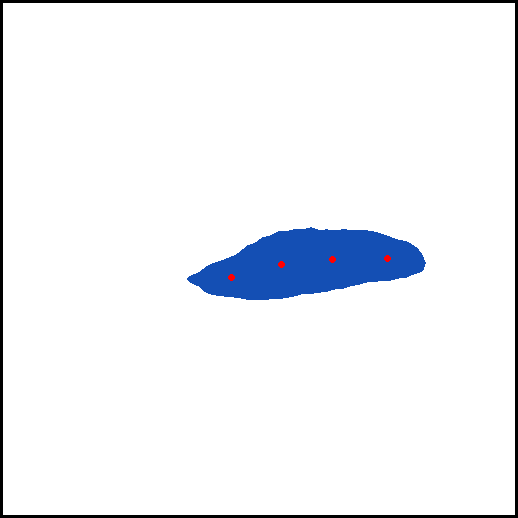}
    \label{fig:A5}
\end{minipage}

\vspace{-11pt}
\begin{minipage}{0.17\textwidth}
\includegraphics[width=\textwidth]{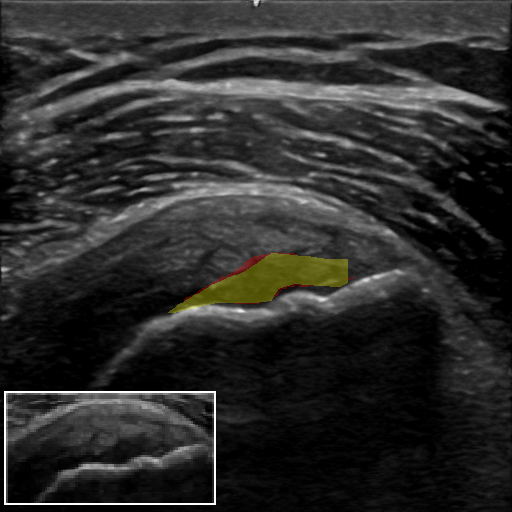}
  \subcaption{100\% of train set.}
  \label{fig:B100}
\endminipage\hfill
\minipage{0.17\textwidth}
\includegraphics[width=\textwidth]{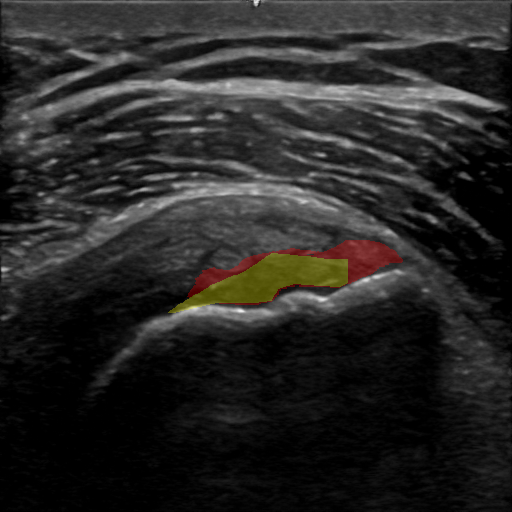}
  \subcaption{35\% of train set.}
    \label{fig:B35}
\endminipage\hfill
\minipage{0.17\textwidth}
\includegraphics[width=\textwidth]{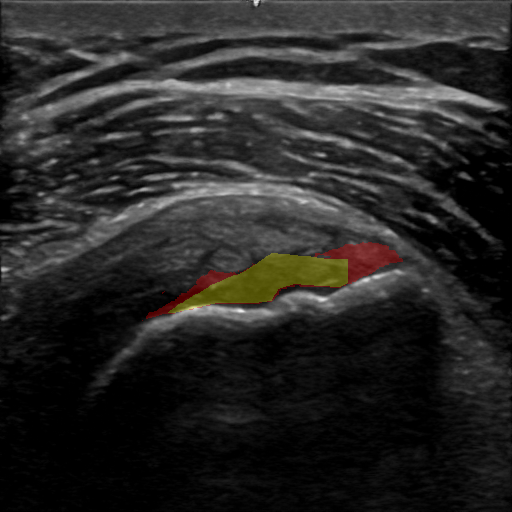}
  \subcaption{15\% of train set.}
    \label{fig:B15}
\endminipage\hfill
\minipage{0.17\textwidth}
\includegraphics[width=\textwidth]{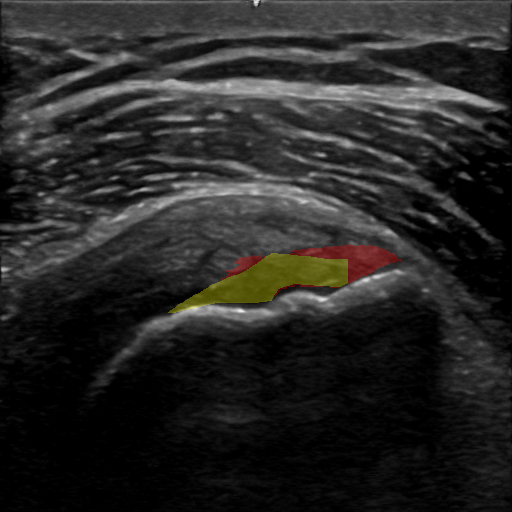}
  \subcaption{8\% of train set.}
  \label{fig:B8}
  \endminipage\hfill
\minipage{0.17\textwidth}
\includegraphics[width=\textwidth]{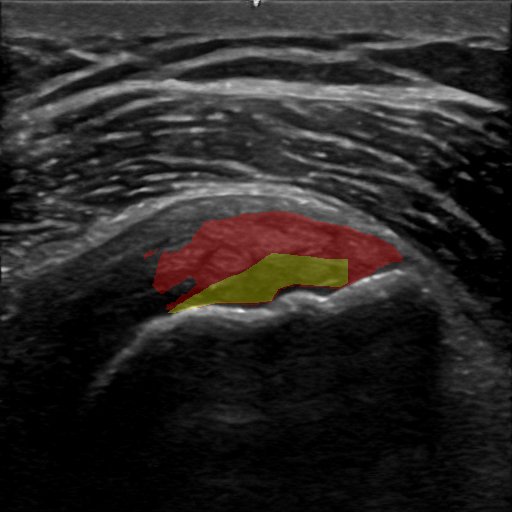}
  \subcaption{5\% of train set.}
    \label{fig:B5}
\end{minipage}
\caption{Positive pathology points retainment in increasingly coarse segmentation mask prediction and our method's results. Top row: Pathology segmentation mask predicted with a DeepLabV3 model trained on varying percent of the training set. Middle row: Positive points selected on binary pathology mask by our positive points selection module.
Bottom row: An illustration of our method's pathology segmentation output,  highlighted in red, compared to the ground-truth segmentation, highlighted in green. The tendon area is shown at the bottom left image for reference. Our method achieves for this test image a Dice similarity coefficient of $0.89, 0.71, 0.73, 0.72, 0.50$ when the coarse segmentor is trained on $100\%,35\%,15\%,8\%,5\%$ of the train set, respectively.}
\label{fig:segmentor_low_data_regimes}
\end{figure*}
\subsection{Dataset}\label{data_section}
The data used for this study is ultrasound images of tendons around the shoulder joint. Specifically, we acquired images of the supraspinatus tendon, infraspinatus tendon, and subscapularis. The images are acquired from both the short-axis and the long-axis views. The main parameters of our data are summarized in Table \ref{tab:dataset_summary}. 

In this work, we aim to segment the partial tear pathology within the tendon, thus our data consists of images paired with the corresponding segmentation mask of anatomies and pathologies. Our data includes semantic labeling of the following classes: DITF, bone, tendon, and muscle. Table \ref{Table:data_semanticlabelling} summarizes the semantic labeling statistics.

In total, our dataset includes 388 images from 124 subjects,  80\% of which are used for training, and the remaining 20\% are used for validation. The test set comprises 40 images.
To prevent data leakage, the test set images are collected from subjects that do not appear in the train data.
All images are resized to a constant resolution of 512x512 pixels.
All data comply with the Institutional Review Board (IRB) data sharing agreement.
\begin{table}[htbp]
\centering
\begin{center}
\caption{Summary of MSK pathology segmentation dataset main parameters.}
\begin{tabular}{p{0.25\linewidth}  p{0.4\linewidth}}
\hline
\textbf{Parameters/Dataset} & \textbf{MSK Ultrasound Images} \\
\hline
\hline
Total frames & 388 \\
Original frame size & 1536 X 796 or 1044 X 646 pixels \\
Subjects & 90 (52.82\% males, 47.18\% females) \\
Average BMI & 24.69 $\pm$ 8.92 \\
Vendor & GE Healthcare\textsuperscript{\texttrademark} \\
Ultrasound system & Logiq S8\textsuperscript{\texttrademark},  Eagle\textsuperscript{\texttrademark},  LogiqE10\textsuperscript{\texttrademark} \\
Data collection & Linear \\
Collection Sites & USA, Israel \\
\hline
\end{tabular}
\label{tab:dataset_summary}
\end{center}
\end{table}

\subsection{Evaluation Metric}
We use the Dice similarity coefficient \cite{Sørensen-Dice} evaluation metric, commonly used in medical image segmentation research to measure the overlapping pixels between prediction and ground truth masks. The Dice similarity coefficient is defined as $\frac{2 \rvert A \cap B \lvert }{\rvert A \lvert + \rvert B \lvert}$, 
where $\mathrm{A}$ and $\mathrm{B}$ are the pixels of the prediction and the ground truth respectively.

\begin{table}[h]
\begin{center}
\caption{Semantic labeling statistics at the 512X512 patches level. M: Million.}
\setlength{\tabcolsep}{5.3pt}
\begin{tabular}{l c c c c} 
 \hline
\textbf{Class} & \textbf{MSK Type} & \makecell{\textbf{Number of images} \\ \textbf{(\% of total)}} & \makecell{\textbf{Total Area} \\\textbf{(pixels)}} & \makecell{\textbf{Mean fraction out}\\ \textbf{of total patch area}} \\ [0.5ex] 
 \hline
   \hline
   Discontinuity in tendon fiber & Pathology & 179 (46.13\%) & 1.11M & 1.09\% \\
   \hline
   Bone & & 288 (74.22\%) & 2.75M & 2.7\% \\
   Tendon & \multirow{2}{*}{Anatomy} & 388 (100\%)  & 10.64M & 10.46\% \\
   Muscle & & 388 (100\%) & 28.13M & 27.65\% \\
 \hline
 \label{Table:data_semanticlabelling}
\end{tabular}
\end{center}
\label{table:dataset_mask_distribution}
\end{table}

\subsection{A Segmentation Model In Low-Data Regimes}
In this experiment, we investigate the performance and properties of a state-of-the-art semantic segmentation model with a limited training set size of MSK ultrasound images.
Our goal is two-fold: \begin{enumerate*}[label=(\roman*)]
\item to validate our conjecture that high-quality prompts can be extracted even from a coarse semantic segmentation prediction, and \item to measure the performance degradation in increasingly low-data regimes. 
\end{enumerate*}
These properties are the basis of our two-stage method for exploiting the advantages of a prompt-able foundation segmentation model. 
Concretely, for an input image $I \in \mathbb{R}^{512 \times 512}$ the segmentation model prediction $S \in \mathbb{R}^{7 \times 512 \times 512} $ corresponds to a semantic segmentation for each class as detailed in Table \ref{Table:data_semanticlabelling}.

\subsection{Segmentation Refinement With Zero-Shot Foundation Models}
\subsubsection*{Positive Points Selection}
A combination of a constant and an adaptive threshold is applied to the coarse segmentation prediction prior to positive point selection. Denote by $c_0$ the coarse segmentation mask prediction at the foreground channel (DITF in our case). We apply a double thresholding mechanism to disregard the noise in the prediction.
\begin{eqnarray}
    \tilde{c} &=& c_0 > t_{min} \\
    c &=& \tilde{c} > 0.4 * \max(\tilde{c})
\end{eqnarray}
The initial threshold screens predictions that lack sufficient global (cross-classes) certainty, when the minimum threshold is set to $t_{min} =0.15$.
The second thresholding term adaptively screens all predictions that lack sufficient local (class-wise) certainty. 
Further, we set the $k$-medoids++ medoid initialization method \cite{K-Means++} which selects more separated initial medoids than those selected by the other methods. The hyper-parameter $k$ is adaptively set such that the sum of distances of samples to their closest cluster center (inertia) is minimized, $k\in [4,6]$. 

\subsubsection*{Negative Points Refinement}
We deploy in our experiments the SonoSAM semantic segmentation foundation model since it is expected to better generalize to zero-shot segmentation of ultrasound images than SAM.

Due to the randomness in the initialization of the complementary positive points $\bar{pts}^{pos}$ selection problem, evaluation is performed over $10$ random initialization.

\subsection{Training Procedure}
Our coarse segmentor is DeepLabV3 \cite{chen2017rethinking}, a state-of-the-art convolutional approach to handle objects in images of varying scales, with a ResNet-50 backbone \cite{he2015deep}. As our complete dataset consists of only 275 training images, the model is pre-trained on the ImageNet dataset \cite{5206848}.
To evaluate our method across different data regimes we trained our coarse segmentor on varying $n$ percentage of the training data, $n \in [5,8,12,20,35,60,100]$, sub-sampled at random. The model is trained with equally weighted BCE loss and a Dice similarity coefficient loss between the predicted and ground-truth segmentation for each class. Each such experiment is trained for 100 epochs, where the weights of the maximal validation loss have been selected for testing.
We used the robust AdamW \cite{loshchilov2017decoupled} optimizer, with no learning rate scheduler and parameters of \(\beta_{1} = 0.9, \beta_{2} = 0.999\) and learning rate of \(4e^{-3}\).
The test set remains constant across the different training experiments.
The model training and evaluation code is implemented with the PyTorch \cite{NEURIPS2019_9015} framework.

\section{Results}\label{results}
\subsection{Semantic Segmentation Model In Low-Data Regimes}
The results of this experiment validate our conjecture that positive pathology points are consistently selected in increasingly coarse segmentation mask predictions. 

As the segmentation model is trained on an increasingly smaller training set, the segmentation mask prediction becomes coarse: the pathology segmentation boundaries become less defined and its prediction probability decreases (Fig. \ref{fig:segmentor_low_data_regimes}, top row). Nevertheless, the positive pathology points selected by our method remain generally consistent (Fig. \ref{fig:segmentor_low_data_regimes}, middle row).   
Consistent with these results, we find that the average Dice similarity coefficient of the segmentation model decreases rapidly when the model is trained on increasingly smaller training set sizes (Fig. \ref{fig:Performance_summary}, `Segmentation Model'). 
These results validate our method's motivation and approach.

\subsection{Semantic Segmentation Refinement With Zero-Shot
Foundation Model}
\begin{figure}[h]
    \centering
    \includegraphics[width=0.45\linewidth]{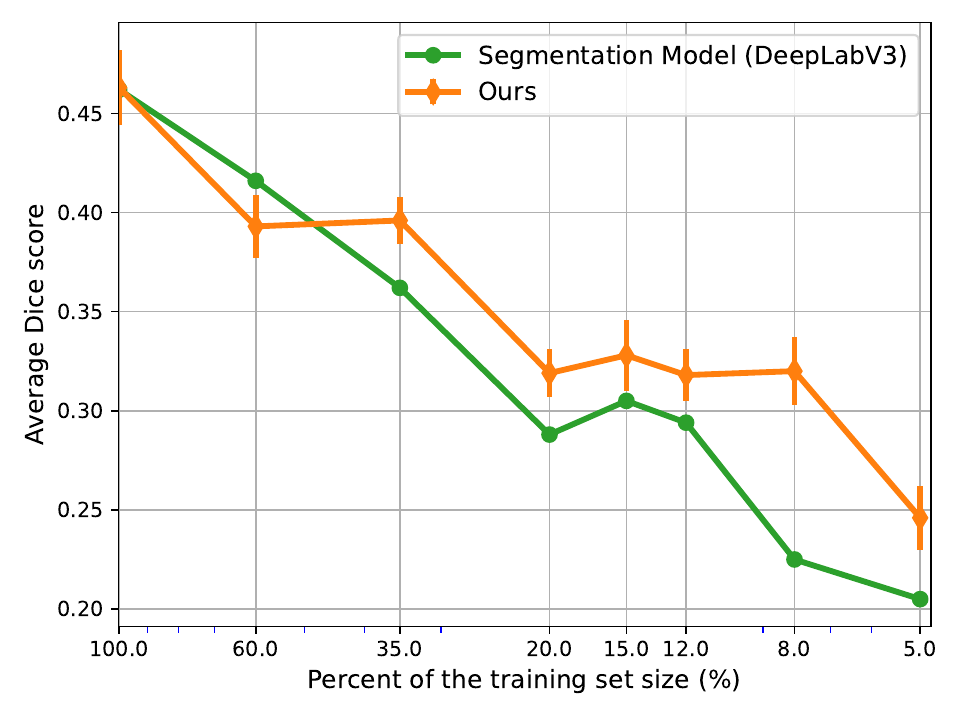}
    \caption{A summary of the average DITF Dice similarity coefficient of methods in various training set sizes.
    Depicted are the results of the baseline segmentation model\cite{chen2017rethinking} and our segmentation refinement with zero-shot SonoSAM foundation model. Error bars depict the standard deviation of our method's statistics. \iffalse A summary of average DITF Dice similarity coefficient of methods in various training set sizes.
    Depicted are results of the baseline segmentation model\cite{chen2017rethinking} and our segmentation refinement with zero-shot SonoSAM foundation model, as well as the results of our method w/o negative points refinement (Sec. \ref{ablation:randomnegatives}), and our segmentation refinement with zero-shot SAM foundation model (Sec. \ref{ablation:ours_w_sam}). The x-axis is in a logarithmic scale.\fi}
    \label{fig:Performance_summary}
\end{figure}

Fig. \ref{fig:Performance_summary} summarizes the results of our method in comparison with those of the baseline segmentation model in
various training set sizes. Our method's average Dice is higher than the baseline's in every training set size. Moreover, our method's performance gain is larger as the training set size decreases ($\sim 10\%$ average Dice increase in 5\% and 8\% training set sizes), substantiating the advantage of our method in low-data regimes.
Our method's pathology segmentation output in
varying training set sizes compared to the ground-truth segmentation is illustrated in Fig. \ref{fig:segmentor_low_data_regimes}, bottom row.

To analyze the stochasticity effect of our method's negative points random initialization (Sec. \ref{negative_selection}), we compare our method's DITF Dice score statistics over ten random initialization and the baseline segmentation model's average DITF Dice similarity coefficient.
Results show that our method's performance is robust, exhibiting relatively low standard deviation in all train set sizes (Fig. \ref{fig:Performance_summary}). Additionally, our method's mean DITF Dice surpasses that of the baseline's in all but one train set size, and is higher by 4\% on average than the baseline. 

\subsection{Ablation Studies}\label{ablation}
In this section, we present ablation studies substantiating the effectiveness of our negative prompt points refinement (NPPR) model, as well as examining our method's performance when replacing the SonoSAM foundation model with SAM. 

\subsubsection{SAM vs. SonoSAM as a segmentation foundation model}\label{ablation:ours_w_sam}
In this study, we investigate the impact of replacing SonoSAM with SAM as the zero-shot semantic segmentation foundation model in our method. 
Table \ref{tbl:ablation_results} shows that harnessing SonoSAM's generalizability for MSK ultrasound images is preferable to SAM in low-data regimes and on par with SAM otherwise.

\subsubsection{Random negative prompt points section}\label{ablation:randomnegatives}
In this experiment, we investigate the effectiveness of our negative prompt points refinement model by comparing it to a random negative prompt points selection algorithm. 
Concretely, $k$ negative prompt points are randomly selected from the modified ground-truth tendon mask, $T^{\tilde{gt}}$.
Our positive points selection approach remains unchanged.
Results in Table \ref{tbl:ablation_results} demonstrate that this naive selection algorithm achieves subpar average Dice scores across almost all train set sizes, especially in low-data regimes.
These results establish the advantage of our negative points optimization algorithm.

\begin{table}[h]
\caption{Ablation studies: quantitative segmentation test results of the mean DITF Dice similarity coefficient (DSC) for different approaches over $10$ run cycles. Our method is using zero-shot SonoSAM \cite{ravishankar2023sonosam} foundation model. A higher DSC is better, with the best scores marked in bold. NPPR: Negative Prompt Points Refinement.}
\begin{center}
\begin{tabular}{l|cccccccc}
\hline
\multirow{2}{*}{\centering \textbf{Methods}} & \multicolumn{8}{c}{\textbf{Percent of the training set}} \\
& \multicolumn{1}{c}{100\%} & \multicolumn{1}{c}{60\%} & \multicolumn{1}{c}{35\%} & \multicolumn{1}{c}{20\%} & \multicolumn{1}{c}{15\%} & \multicolumn{1}{c}{12\%} & \multicolumn{1}{c}{8\%} & \multicolumn{1}{c}{5\%}\\
\hline
\hline
Ours without NPPR & 44.6\% & 40.0\% & 34.2\% & 27.8\% & 30.3\% & 27.5\% & 20.7\% & 16.6\%\\
\hline
Ours with SAM & 45.5\% & \textbf{41.6\%} & \textbf{39.7\%} & 29.3\% & \textbf{32.9\%} & 28.3\% & 27.6\% & 23.0\%\\
Ours & \textbf{46.3}\% & 39.3\% & 39.6\% & \textbf{31.9\%} & 32.8\% & \textbf{31.8\%} & \textbf{32.0\%} & \textbf{24.6\%}\\
\hline

\end{tabular}
\end{center}
\label{tbl:ablation_results}
\end{table}

\section{Conclusions}\label{conclusions}
In this paper, we address the performance degradation of a state-of-the-art semantic segmentation model in low-data regimes. A novel prompt points selection algorithm optimized on a zero-shot segmentation foundation model was presented, as a means of refining a coarse pathology segmentation. 

Our method's advantages are brought to light in varying degrees of low-data regimes experiments, demonstrating a larger performance gain compared to the baseline segmentation model as the training set size decreases (Fig. \ref{fig:Performance_summary}). Further, we validate our method's robustness to negative point initialization stochasticity and study the effectiveness of our prompt points refinement model (Section \ref{ablation:randomnegatives}). Results demonstrate that the generalization of SonoSAM in extremely low data regimes is better than SAM's (Section \ref{ablation:ours_w_sam}). Our approach can be used for other ultrasound-based medical diagnostics tasks.
An inherent limitation of our two-stage method is that its latency is higher than that of a core segmentation model.

%\bibliography{bib}
%\bibliographystyle{plainnat}

\bibliographystyle{unsrt}

\end{document}